# Mixed-Methods Exploration of Sociotechnical Factors to Improve Adoption of Crime-Reporting Apps in Urban Communities

Olalekan David-Banjo, Chinazunwa Uwaoma, Paul Witman

Crime-reporting mobile applications can enhance community safety and engagement in urban areas, yet their adoption remains low, partly because developers prioritize technological features over residents' social realities. This study used an exploratory sequential mixed-methods design to identify and validate the sociotechnical factors that drive adoption. Two focus groups and an analysis of 660 app reviews revealed that users' initial sense of usefulness fades, whereas sustained engagement depends on perceived value (PV), comprising six dimensions: safety, anonymity, recency, autonomy, camaraderie, and accessibility. A survey of 505 urban residents, analyzed with PLS-SEM, validated PV as a formative higher-order construct formed by four of these dimensions (safety, anonymity, autonomy, and camaraderie). PV was the strongest predictor of behavioral intention, ahead of effort expectancy, facilitating conditions, and performance expectancy, and behavioral intention strongly predicted actual use. The model explained 75% of the variance in behavioral intention and 51% in actual use. The findings offer design principles for community-centered crime-reporting technologies.

**Keywords:** crime-reporting apps; sociotechnical factors; mixed-methods; PLS-SEM; perceived value; technology adoption.

## INTRODUCTION

Although crime-reporting mobile applications (apps) can enhance safety and foster social cohesion, their persistently low use in urban communities limits their societal impact and excludes vulnerable populations from important safety resources. Developers often prioritize technological sophistication over community needs, resulting in solutions that are misaligned with the technological frames of urban residents (Orlikowski & Gash, 1994; Maedche et al., 2019). The issue is pressing: the National Crime Victimization Survey reports that urban violent victimization remained elevated at 29.6 per 1,000 persons in 2023, while urban property victimization rose from 176.1 to 192.3 per 1,000 households between 2022 and 2023 (Tapp & Coen, 2024).

We applied an exploratory sequential mixed-methods design to investigate sociotechnical factors influencing crime-reporting app adoption, guided by three questions: (1) How do urban residents perceive the social value of crime-reporting apps, and what social factors influence their willingness to use them? (2) How do urban residents perceive an app's technological features, and which features matter most to them? (3) From a user perspective, what sociotechnical factors are crucial to app design? This research advances understanding and has practical design implications for community safety technologies.

## BACKGROUND

Sociotechnical systems theory highlights the interplay between technical systems and the social contexts in which they are embedded (Baxter & Sommerville, 2011). However, software developers frequently design crime-reporting apps in isolation, paying little attention to the contexts in which these technologies operate. Consequently, designs are often dominated by designers' technological frames, thus misaligning with community needs and creating adoption challenges (Orlikowski & Gash, 1994). The conflicting needs, goals, and requirements of stakeholders exacerbate this misalignment (Maedche et al., 2019).

Orlikowski and Iacono (2001) critique research that takes the IT artifact for granted and treats technologies as "relatively stable, discrete, independent, and fixed" (p. 121). This reductionist view ignores the social and organizational dynamics that ultimately determine how technologies are interpreted, adapted, and sustained in practice (Orlikowski & Gash, 1994). Similarly, Benbasat and Zmud (2003) emphasize that IT developments are inseparable from the "structures, routines, norms, and values" (p. 186) of the environments in which they are instantiated.

Although participatory design research suggests that artifacts can be co-created with end users (Kensing & Blomberg, 1998; Sanders & Stappers, 2008), developers of sociotechnical systems often overlook social or community values as design considerations, as documented in related domains (Dillahunt et al., 2018; Ehsan et al., 2023). This oversight often produces unintended outcomes (Whitworth, 2011).

This study addresses these issues by integrating qualitative insights from urban residents with quantitative validation using partial least squares structural equation modeling (PLS-SEM). It examines perceived value (PV) as a multifaceted sociotechnical construct. Six dimensions emerged qualitatively: sense of safety, anonymity, recency, autonomy, camaraderie, and accessibility; four of these were retained in the validated model. To predict adoption, PV was assessed alongside constructs of the unified theory of acceptance and use of technology (UTAUT; Venkatesh et al., 2003). The study's theoretical contribution extends technology adoption models with user-defined constructs, offering actionable insights for designing trustworthy, relevant crime-reporting tools.

## METHODOLOGY

An exploratory sequential mixed-methods design (Creswell & Plano Clark, 2017) was applied (Fig. 1). In phase 1, two virtual focus groups were conducted with urban residents of Orange County, California (n=5; n=6), recruited via UserInterviews.com. Participants were adult smartphone owners with prior crime-reporting app experience. Discussions followed a semi-structured protocol and were complemented by content analysis of 660 Google Play reviews of the crime-reporting apps Citizen (n=209) and SpotCrime (n=451).

In phase 2, an online Zoho survey of residents of densely populated U.S. cities yielded 505 valid responses from 748 submitted questionnaires (67.5%); 53.1% were women, 54.1% were aged 40 or older, and 47.5% held undergraduate degrees. Items used five-point Likert scales. UTAUT items were adapted from Venkatesh et al. (2003), and PV items were developed from phase 1 insights (Tables 1 and 2).

PV was specified as a second-order formative construct formed from reflective first-order dimensions and estimated using the disjoint two-stage approach (Sarstedt et al., 2019). Following UTAUT2 (Venkatesh et al., 2012), facilitating conditions predicted behavioral intention. Six UTAUT indicators (PE2, PE4, FC3, FC4, SI1, SI2) and two PV dimensions (recency, accessibility) were removed owing to discriminant validity violations; consequently, social influence was excluded from the final model.

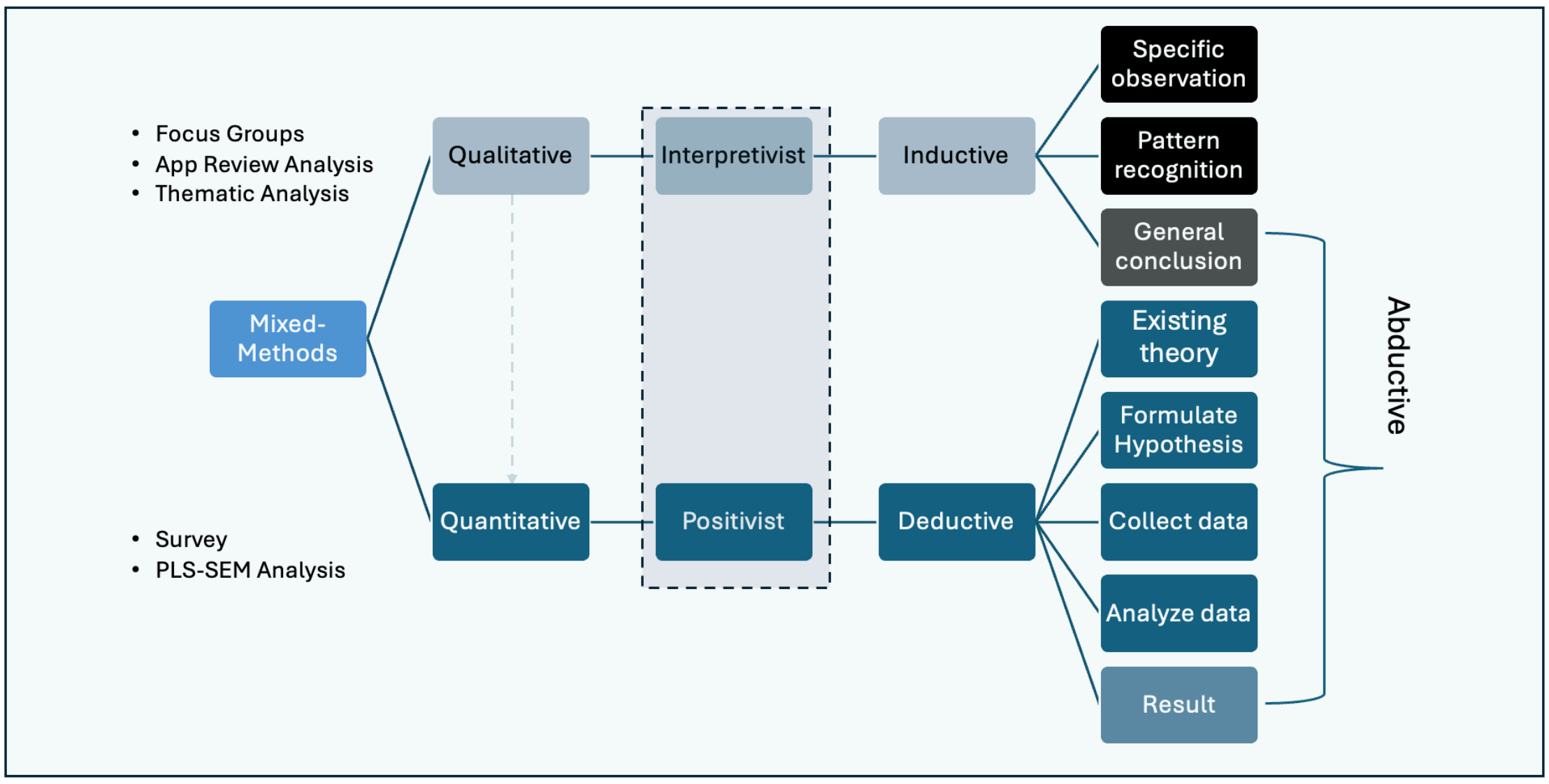

Fig. 1. Exploratory sequential mixed-methods design

**Table 1.** Measurement items for perceived value dimensions and item retention

| Construct | Indicators | Questions | Retained in model? (Yes or No) |
|---|---|---|---|
| Sense of safety | SS1 | The notifications and incident reports I receive from the crime-reporting app make me feel safe. | Yes |
| | SS2 | Using the crime-reporting app makes me feel less likely to be targeted by crime. | Yes |
| | SS3 | The crime-reporting mobile app empowers me to recognize and respond appropriately to suspicious activities. | Yes |
| Sense of autonomy | SAU1 | I decide how and when to use the crime-reporting app. | Yes |
| | SAU2 | When using the crime-reporting mobile app, I can customize the notifications and incident radius to suit my needs. | Yes |
| Sense of anonymity | SAN1 | I am confident that other users cannot identify me through the app. | Yes |
| | SAN2 | When I report an incident, my identity remains protected and anonymous. | Yes |
| Sense of recency | SR1 | The crime-reporting mobile app provides timely and accurate crime updates. | No |
| | SR2 | The crime-reporting app keeps me informed about the latest incidents in my neighborhood. | No |
| Sense of camaraderie | SC1 | When using the crime-reporting mobile app, I feel a sense of shared responsibility with others. | Yes |
| | SC2 | I am more likely to use the crime-reporting mobile app if others in my neighborhood use it as well. | Yes |
| | SC3 | I enjoy engaging with other users on the crime-reporting app platform. | Yes |
| Sense of accessibility | SAC1 | The crime-reporting mobile app is accessible to people with little experience. | No |
| | SAC2 | The crime-reporting mobile app's design accommodates users with different abilities and backgrounds. | No |

**Table 2.** Measurement items, sources, and retention status

| Construct | | Indicators | Survey items and codes | Sources | Retained in model? (Y(es) or N(o)) |
|---|---|---|---|---|---|
| Performance expectancy | Perception that the system will improve task performance | PE1 | Using the crime-reporting mobile app makes it easier for me to report crimes or suspicious activities. | Davis, 1989; Davis et al., 1989, as cited in Venkatesh et al., 2003 | Yes |
| | | PE2 | This crime-reporting mobile app helps me quickly notify people about incidents in my area. | | No |
| | | PE3 | By using this app, I can contribute to making my community safer. | | Yes |
| | | PE4 | The crime-reporting mobile app is useful for staying informed about crime and safety issues nearby. | | No |
| Effort expectancy | Degree of ease associated with system use | EE1 | Learning to use the crime-reporting mobile app is easy for me. | Davis, 1989; Davis et al., 1989, as cited in Venkatesh et al., 2003 | Yes |
| | | EE2 | It is easy to report incidents using the crime-reporting mobile app. | | Yes |
| | | EE3 | It is easy for me to navigate the crime-reporting mobile app and find the features I need. | | Yes |
| | | EE4 | The instructions provided in the crime-reporting mobile app are clear and understandable. | | Yes |
| Facilitating conditions | Extent to which the individual believes that technical infrastructure exists to support system use | FC1 | I have access to a smartphone and an internet connection to use the crime-reporting mobile app. | Thompson et al. 1991, as cited in Venkatesh et al. 2003 | Yes |
| | | FC2 | I have the knowledge necessary to use the crime-reporting mobile app. | | Yes |
| | | FC3 | If I encounter any issues while using the crime-reporting mobile app, I can easily get help or support. | | No |
| | | FC4 | The crime-reporting mobile app works well with other features or apps on my phone (e.g., camera, GPS). | | No |
| Social Influence | Belief that people important to the user think they should use the app | SI1 | People in my community think I should use this app for reporting crime. | Venkatesh et al. (2003); Fishbein & Ajzen (1975) | No |
| | | SI2 | I would use the app if many of my friends or neighbors recommended it. | | No |
| Behavioral intention | Individual's intention to use the system | BI1 | I am likely to recommend the crime-reporting mobile app to others in my community. | Davis et al., 1989, as cited in Venkatesh et al. 2003 | Yes |
| | | BI2 | I will regularly use the crime-reporting mobile app when I need to report an incident. | | Yes |
| | | BI3 | I intend to use the crime-reporting mobile app to report | | Yes |
| | | | crimes or suspicious activities in the future. | | |
| Actual use | Extent of system use | AU1 | I use the crime-reporting mobile app. | Venkatesh et al., 2000, as cited in Venkatesh et al. 2003 | Yes |

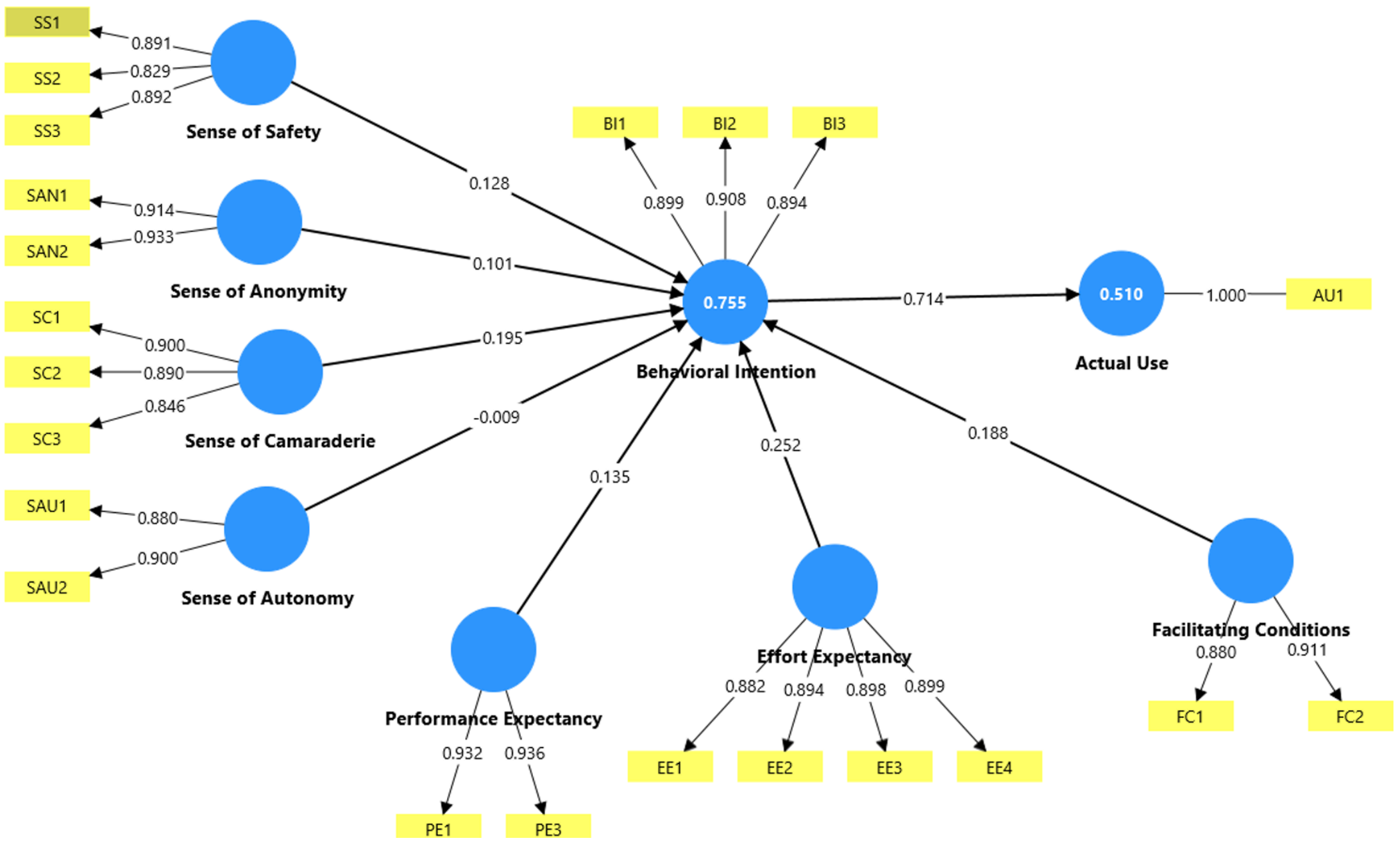

Fig. 2. First-order reflective measurement model

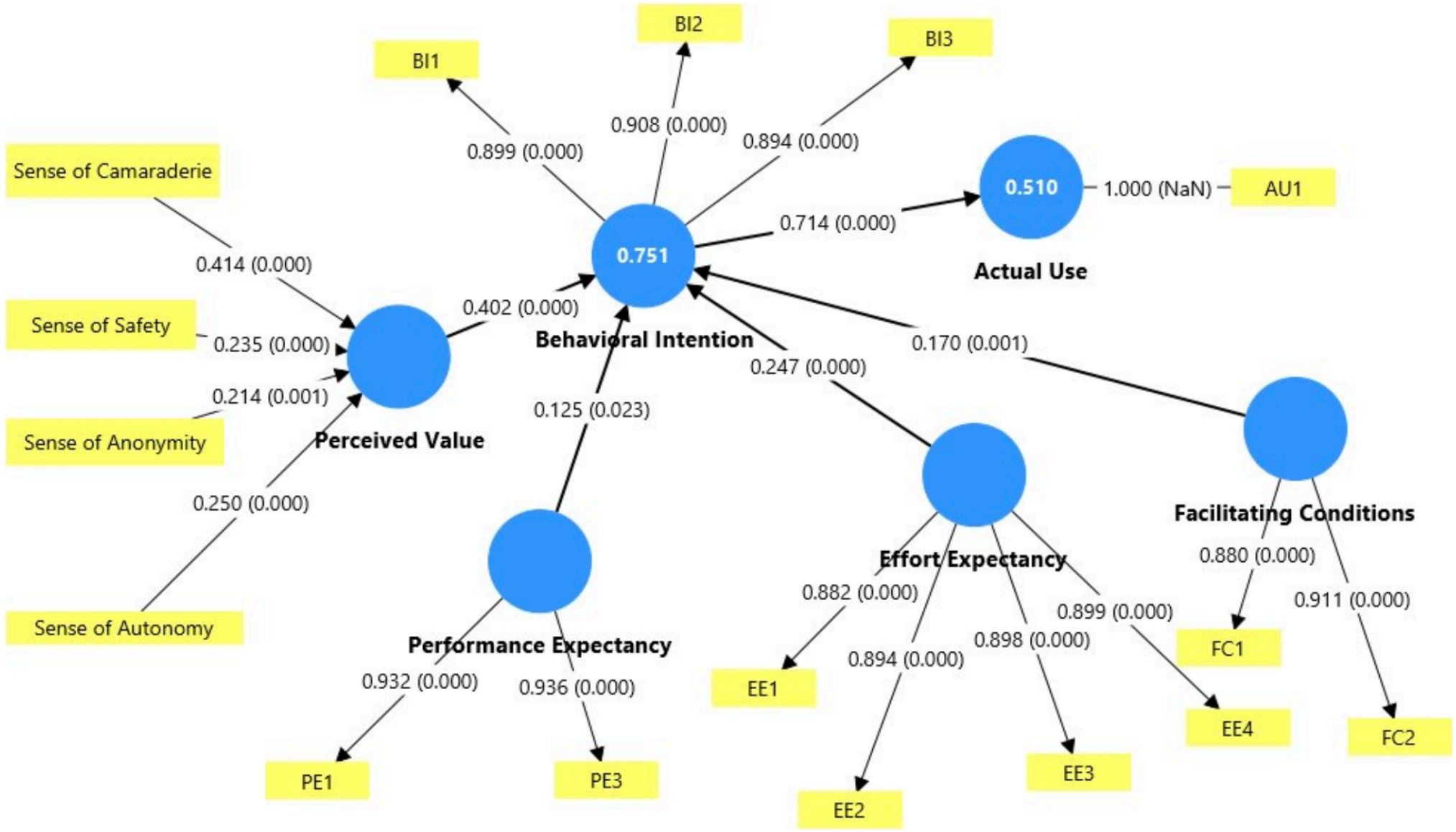

Fig. 3. Second-order perceived value model
Notes: Values in parentheses are p-values. AU1 is a single-item measure (loading fixed at 1.000).

## RESULTS AND DISCUSSION

### Phase 1: Qualitative results

Thematic analysis of the focus group transcripts revealed two central themes. First, transient usefulness, the initial impression that the app served its purpose, waned when the app failed to meet evolving needs. Second, PV, a more enduring, context-driven judgment of the app's relevance, was shaped by app features and users' social realities. Six PV dimensions emerged.

*Sense of safety* was foundational. Participants expected reassurance, vigilance, and support for preventive action (Farrall et al., 2000). *Sense of anonymity* was a precondition for reporting without fear of retaliation, consistent with greater self-disclosure under anonymity (Joinson, 2001; Suler, 2004). *Sense of recency*, the timeliness and accuracy of incident information, underpinned credibility, in line with DeLone and McLean's (2003) emphasis on information quality. *Sense of autonomy* reflected users' need to control interactions by filtering content and adjusting notifications; users with such control typically engage proactively (Spreitzer, 1995; Deci & Ryan, 2000). *Sense of camaraderie* captured expectations of connectedness and collective vigilance, consistent with Putnam's (1995) work on social capital. *Sense of accessibility* emphasized usability across devices, abilities, and languages.

A directed content analysis (Hsieh & Shannon, 2005) of app reviews largely reinforced these dimensions. Recency, accessibility, safety, and autonomy were most frequently mentioned, while anonymity (6 of 660) and camaraderie (14) were rare in reviews but salient in the focus groups (Table 3).

**Table 3.** Comparison of thematic mentions between Citizen and SpotCrime

| Theme | Citizen (n=209) | SpotCrime (n=451) | Total mentions |
|---|---|---|---|
| Sense of recency | 48 | 114 | 162 |
| Sense of anonymity | 0 | 6 | 6 |
| Sense of autonomy | 65 | 71 | 136 |
| Sense of camaraderie | 1 | 13 | 14 |
| Sense of safety | 21 | 124 | 145 |
| Sense of accessibility | 60 | 86 | 146 |
| Other | 14 | 37 | 51 |

### Phase 2: PLS-SEM results

*Measurement model assessment*

The reflective measurement model (Fig. 2) showed adequate reliability and validity: loadings exceeded 0.708 (0.829–0.936), Cronbach's α exceeded 0.70 (0.739–0.916), composite reliability remained below 0.95 (0.885–0.940), and AVE exceeded 0.50 (0.759–0.872) (Table 4) (Hair et al., 2019). No HTMT bootstrapped confidence interval included 1.0 (Henseler et al., 2015), although several values approached or exceeded 0.90 (e.g., performance expectancy–effort expectancy = 0.908), so results involving these constructs warrant caution (Franke & Sarstedt, 2019) (Table 5).

For the formative higher-order PV construct, collinearity was not a concern (VIF = 2.469–3.425). All outer weights were positive and significant ($p \leq .001$): camaraderie ($w = 0.414$), autonomy (0.250), safety (0.235), and anonymity (0.214). All outer loadings exceeded 0.86 ($p < .001$) (Table 6).

**Table 4.** Reliability and validity results

| Constructs | Indicators | Loadings >0.7 | **Cα >0.7** | **rho_A >0.7** | **CR >0.7** | **AVE >0.5** |
|---|---|---|---|---|---|---|
| Actual use | | | | | | |
| | AU1 | 1.000 | | | | |
| Behavioral intention | | | 0.883 | 0.884 | 0.928 | 0.810 |
| | BI1 | 0.899 | | | | |
| | BI2 | 0.908 | | | | |
| | BI3 | 0.894 | | | | |
| Performance expectancy | | | 0.853 | 0.854 | 0.932 | 0.872 |
| | PE1 | 0.932 | | | | |
| | PE3 | 0.936 | | | | |
| Effort expectancy | | | 0.916 | 0.917 | 0.940 | 0.798 |
| | EE1 | 0.882 | | | | |
| | EE2 | 0.894 | | | | |
| | EE3 | 0.898 | | | | |
| | EE4 | 0.899 | | | | |
| Facilitating conditions | | | 0.754 | 0.765 | 0.890 | 0.802 |
| | FC1 | 0.880 | | | | |
| | FC2 | 0.911 | | | | |
| Sense of safety | | | 0.841 | 0.853 | 0.904 | 0.759 |
| | SS1 | 0.891 | | | | |
| | SS2 | 0.829 | | | | |
| | SS3 | 0.892 | | | | |
| Sense of anonymity | | | 0.828 | 0.836 | 0.920 | 0.853 |
| | SAN1 | 0.914 | | | | |
| | SAN2 | 0.933 | | | | |
| Sense of camaraderie | | | 0.853 | 0.856 | 0.911 | 0.773 |
| | SC1 | 0.900 | | | | |
| | SC2 | 0.890 | | | | |
| | SC3 | 0.846 | | | | |
| Sense of autonomy | | | 0.739 | 0.743 | 0.885 | 0.793 |
| | SAU1 | 0.880 | | | | |
| | SAU2 | 0.900 | | | | |

Notes: Actual use is a single-item construct; reliability and AVE are not applicable.

**Table 5**. Discriminant validity (bootstrapping): HTMT inference

| | Original sample (O) | Sample mean (M) | Bias | 5.0% | 95.0% |
|---|---|---|---|---|---|
| Behavioral intention ↔ actual use | 0.759 | 0.759 | -0.000 | 0.712 | 0.802 |
| Effort expectancy ↔ actual use | 0.673 | 0.673 | -0.001 | 0.615 | 0.725 |
| Effort expectancy ↔ behavioral intention | 0.894 | 0.894 | -0.000 | 0.857 | 0.927 |
| Facilitating conditions ↔ actual use | 0.609 | 0.608 | -0.000 | 0.540 | 0.674 |
| Facilitating conditions ↔ behavioral intention | 0.880 | 0.880 | -0.000 | 0.816 | 0.936 |
| Facilitating conditions ↔ effort expectancy | 0.890 | 0.891 | 0.001 | 0.822 | 0.948 |
| Performance expectancy ↔ actual use | 0.613 | 0.612 | -0.001 | 0.546 | 0.674 |
| Performance expectancy ↔ behavioral intention | 0.891 | 0.891 | -0.000 | 0.847 | 0.930 |
| Performance Expectancy ↔ effort expectancy | 0.908 | 0.907 | -0.000 | 0.870 | 0.939 |
| Performance expectancy ↔ facilitating conditions | 0.882 | 0.882 | 0.000 | 0.815 | 0.941 |

*Structural model assessment*

The model explained substantial variance in behavioral intention ($R^2 = 0.751$) and moderate variance in actual use ($R^2 = 0.51$). Inner VIF values (2.445–4.156; Table 7) indicated no critical collinearity, although values above 3 call for cautious interpretation (Hair et al., 2019). PV was the strongest predictor of behavioral intention ($\beta = 0.402$, $p < .001$), exceeding effort expectancy ($\beta = 0.247$, $p < .001$), facilitating conditions ($\beta = 0.170$, $p = .001$), and performance expectancy ($\beta = 0.125$, $p = .023$). Behavioral intention strongly predicted actual use ($\beta = 0.714$, $p < .001$) (Table 8). PLSpredict yielded $Q^2$predict > 0 for all indicators. Behavioral intention indicators outperformed the linear model benchmark, indicating high predictive power, whereas actual use did not (RMSE = 1.014 vs. 0.997), indicating no predictive power for that single item (Shmueli et al., 2019) (Table 9).

**Table 6.** Outer weights and outer loadings of perceived value dimensions: bootstrapping results

| Dimension → perceived value | Outer weights | | | | Outer loadings | | |
|---|---|---|---|---|---|---|---|
| | **O** | **STDEV** | ***t*** | ***p*** | **O** | ***t*** | ***p*** |
| Sense of anonymity | 0.214 | 0.064 | 3.345 | 0.001 | 0.869 | 37.129 | 0.000 |
| Sense of autonomy | 0.250 | 0.059 | 4.223 | 0.000 | 0.862 | 34.435 | 0.000 |
| Sense of camaraderie | 0.414 | 0.078 | 5.280 | 0.000 | 0.940 | 59.601 | 0.000 |
| Sense of safety | 0.235 | 0.067 | 3.504 | 0.000 | 0.888 | 35.655 | 0.000 |

Notes: Stage 2 of the disjoint two-stage approach; bootstrapping results. O = original sample; STDEV = standard deviation.

**Table 7.** Collinearity assessment (inner model)

| | VIF |
|---|---|
| Effort expectancy → behavioral intention | 4.156 |
| Facilitating conditions → behavioral intention | 2.445 |
| Perceived value → behavioral intention | 4.068 |
| Performance expectancy → behavioral intention | 3.773 |

**Table 8.** Significance and relevance of path coefficients

| | Original sample (O) | Sample mean (M) | Standard deviation (STDEV) | T statistics (\|O/STDEV\|) | f-square | P values |
|---|---|---|---|---|---|---|
| Behavioral intention → actual use | 0.714 | 0.714 | 0.027 | 26.194 | 1.041 | 0.000 |
| Effort expectancy → behavioral intention | 0.247 | 0.246 | 0.063 | 3.899 | 0.059 | 0.000 |
| Facilitating conditions → behavioral intention | 0.170 | 0.169 | 0.053 | 3.214 | 0.048 | 0.001 |
| Perceived value → behavioral intention | 0.402 | 0.407 | 0.059 | 6.754 | 0.159 | 0.000 |
| Performance expectancy → behavioral intention | 0.125 | 0.123 | 0.055 | 2.266 | 0.017 | 0.023 |

**Table 9.** PLSpredict MV summary

| | $Q^2$predict | PLS-SEM_RMSE | PLS-SEM_MAE | LM_RMSE | LM_MAE | IA_RMSE | IA_MAE |
|---|---|---|---|---|---|---|---|
| BI1 | 0.647 | 0.642 | 0.459 | 0.651 | 0.462 | 1.081 | 0.841 |
| BI2 | 0.570 | 0.734 | 0.511 | 0.735 | 0.513 | 1.120 | 0.882 |
| BI3 | 0.578 | 0.708 | 0.495 | 0.724 | 0.504 | 1.090 | 0.834 |
| AU1 | 0.469 | 1.014 | 0.765 | 0.997 | 0.723 | 1.392 | 1.173 |

## CONCLUSIONS AND IMPLICATIONS FOR THEORY AND PRACTICE

This study extends technology adoption research on community-based artifacts by showing that PV, formed by senses of safety, anonymity, autonomy, and camaraderie, is the strongest driver of intention to adopt crime-reporting apps. Theoretically, it extends UTAUT with community-defined sociotechnical factors and demonstrates the importance of aligning social and technical systems in community safety innovative design.

Practically, developers, civic technologists, and public safety agencies should design crime-reporting apps that prioritize community connectedness, user control, identity protection, and intuitive interfaces. Qualitative findings also suggest that timely and accurate incident updates matter to users. These transferable design principles can improve adoption, trust, and sustained engagement.

This study's limitations include cross-sectional, self-reported data, including actual use, raising the possibility of common method bias, and PV dimensions derived from a small single-county sample; recency and accessibility could not be validated, and social influence was excluded. Future research should examine longitudinal usage dynamics, use refined measures to retest the excluded dimensions, and evaluate feature-specific interventions that support community-centered safety outcomes.